\documentclass[letterpaper, 10 pt, conference]{ieeeconf}  

\IEEEoverridecommandlockouts                              

\usepackage{graphicx} 
\usepackage{subfigure}
\usepackage{multirow}
\usepackage{cite}
\usepackage{siunitx}
\usepackage[T1]{fontenc}
\title{\LARGE \bf
Navigating to Objects in Unseen Environments by Distance Prediction
}

\author{Minzhao Zhu, Binglei Zhao, Tao Kong 
\thanks{Minzhao Zhu and Binglei Zhao contribute equally. ByteDance AI Lab, Beijing, China. {\tt\small \{zhaobinglei, zhuminzhao, kongtao\}@bytedance.com}
}
}%

\begin{document}

\maketitle
\thispagestyle{empty}
\pagestyle{empty}

\begin{abstract}
Object Goal Navigation (ObjectNav) task is to navigate an agent to an object category in unseen environments without a pre-built map. 
In this paper, we solve this task by predicting the distance to the target using semantically-related objects as cues. Based on the estimated distance to the target object, our method directly choose optimal mid-term goals that are more likely to have a shorter path to the target.
Specifically, based on the learned knowledge, our model takes a bird's-eye view semantic map as input, and estimates the path length from the frontier map cells to the target object. With the estimated distance map, the agent could simultaneously explore the environment and navigate to the target objects based on a simple human-designed strategy.
Empirical results in visually realistic simulation environments show that the proposed method outperforms a wide range of baselines on success rate and efficiency. 
Real-robot experiment also demonstrates that our method generalizes well to the real world.


\end{abstract}

\section{INTRODUCTION}
Object Goal Navigation (ObjectNav) \cite{objectnav_defination} is one of the fundamental embodied navigation tasks. In this task, an intelligent agent is required to move to the location of a target object category in an unseen indoor environment. In classical navigation tasks, under normal circumstances, the map of the environment is constructed in advance. Therefore, a goal location can be given to the agent in the form of coordinates on that map. 
However, in the ObjectNav task, a pre-built map is unavailable, and the goal coordinate is not given. 
Therefore, the agent has to explore the environment and search for the target object. 

During searching, there are many possible areas to be explored. How to prioritize these candidate goals in order to improve exploration efficiency? Obviously, in a new environment, the only information we can use is the knowledge we have learned in other similar environments, such as the spatial relations between objects.
With this commonsense, humans tend to explore the object that is usually close to the target object.
For example, if our target is a chair, we should explore around a table first, while temporarily skipping other regions.
This is because we know chairs are often adjacent to tables; thus, if we move toward the table, it is more likely to find a chair in that direction than other directions away from that table. If we incorporate this kind of prior knowledge into a spatial map, we can transform the ObjectNav task into a classical navigation problem.

\begin{figure}[thpb]
    \centering
    \includegraphics[width=0.37\textwidth]{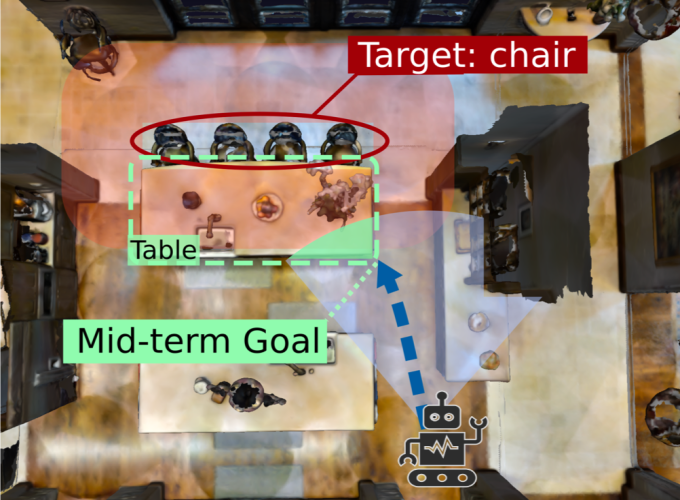}
    \caption{In Object Goal Navigation (ObjectNav) task, an agent is required to navigate to the location of a target object category in an unseen environment using an RGB-D sensor.
    We propose to solve this task by predicting the distance to the target based on spatial relations between objects. The blue shadow in the figure indicates the agent's field of view. The red shadow indicates the area around the target object. In this example, the target object is "chair." The distance to "chair" is supposed to be estimated based on the observed object "table." Since the distance from "table" to "chair"  is often smaller than from other objects, navigating toward "table" are more likely to find "chair" compared with other directions. 
    }
    \label{fig: topic_illustrate}
    \vspace{-0.3cm}
\end{figure}

Inspired by this, we propose to solve the ObjectNav task by predicting the distance to the target using semantically-related objects as cues. Inspired by the concept of Frontier-based Exploration~\cite{frontier_based_exploration}, we predict the distance value (the length of the shortest path to the nearest target object) on the frontier area, and then select goals on the frontier according to the predicted distance.

The Target Distance Prediction model, the core module in our system, learns to predict the distance to the target based on the explored semantic map. 
The model is encouraged to capture the spatial relations between different semantics. For example, if the target object is 'chair,' the predicted distance around tables should be small.

Many recent works~\cite{baseline, attention, 2021winner, sscnav, transformer1} formulate this task as Reinforcement Learning (RL) and have achieved good performance. However, RL often suffers from low sample efficiency~\cite{sample_efficiency}. In contrast, we formulate it as a target distance estimation problem, which could be optimized in a fully supervised way during training. Our method does not require interacting with the environment to get a reward during training.
There are also recent works~\cite{sscnav, map_predict} training a semantic scene completion model to learn the prior knowledge.
However, it is difficult, even for humans, to predict the exact location of related objects. For example, although we know chairs may be close to tables, we cannot predict the chair's relative pose to the table accurately, since the chair could be placed anywhere around that table. In contrast, the distance between the chairs and tables does not vary too much. Our method only needs to predict this distance, which is relatively easier to learn.

Our main contribution is that we formulate the ObjectNav task as a target distance prediction problem based on spatial relations between objects. Our method consists of three parts. First, given the RGB-D image and the agent's pose, a bird's-eye view semantic map is incrementally built. Then, based on the semantic map, a target distance map is predicted. The cells in this distance map store the predicted path length to the nearest target object. Then the distance map is fed to the local policy to get action.
Since the model's output is the estimated distance to the target, a mid-term goal can be easily selected in order to get the shortest path to the target.
As a result, it can be easily integrated into a traditional navigation system.
In this paper, we use several simple goal-select strategies and path planning algorithms to show the effectiveness of the distance map.
We perform experiments in the Matterport3D dataset using the Habitat simulator. Our method outperforms the baseline~\cite{baseline} method with an improvement of 2.6\% success rate and 0.035 SPL(Success weighted by normalized inverse Path Length)~\cite{SPL_defination}. The Experiment on our robot base also demonstrates that our method generalizes well to the real world. 


\begin{figure*}[thpb]
    \centering
    \vspace{0.15cm}
    \includegraphics[width=0.96\textwidth]{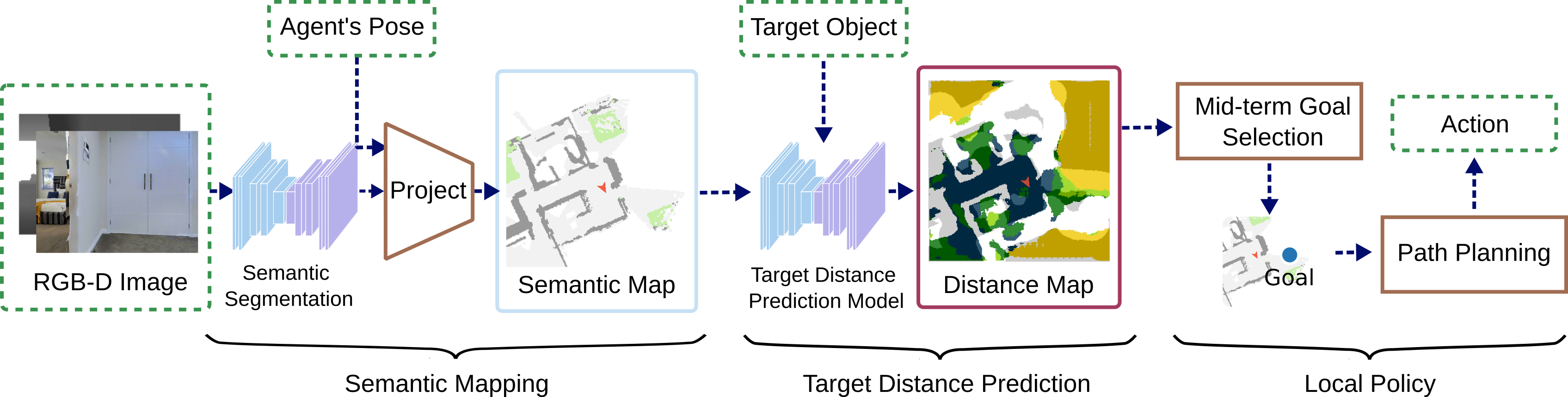}
    \caption{\textbf{System Overview.} 
    Our method consists of a semantic mapping module, a target distance prediction module, and a local policy. Given the RGB-D observation and the agent's pose, the \textbf{Semantic Mapping Module} builds a bird's-eye view semantic map. Then, the \textbf{Target Distance Prediction Module} predicts the distance to the target object on the cells around the exploration frontiers. Based on the distance map, the \textbf{Local Policy} chooses a mid-term goal and gets action. }
    \label{fig: system_overview}
    \vspace{-0.3cm}
\end{figure*}

\section{RELATED WORK}

\subsection{Active SLAM}
Given a specific goal location, the classical navigation methods\cite{probabilistic_robotics} focus on path planning\cite{FMM,a*, RRT, riskmap} based on a previously constructed map.
However, Active SLAM\cite{activeslam1} aims to explore unknown experiments and build maps automatically, which is a decision-making problem tightly coupled with localization, mapping, and planning.
Frontier-based Exploration~\cite{frontier_based_exploration} selects mid-term goals on the frontier according to a designed strategy and extends its map by moving to new frontiers. 
Some methods~\cite{pomdp2} formulate this problem as a POMDP (Partially Observable Markov Decision Process)~\cite{pomdp}.
Recently, some researchers have designed learning-based policies\cite{act_slam_policiy1, act_slam_policiy2, act_slam_policiy8,act_slam_policiy9} to tackle this problem. 
Chaplot et al.\cite{act_slam_policiy2} propose a novel module that combines a hierarchical network design and classical path planning, which significantly improves the sample efficiency and leads to better performance.
The main aim of Active SLAM is to explore the environment efficiently, while the ObjectNav task requires the agent to search the target object category efficiently.

\subsection{Object Goal Navigation}
While Active SLAM is to explore the environment efficiently, Target Driven Navigation is to navigate to a given target in an unknown environment~\cite{target_rl}. The goal could be a point, an object category, a room, etc.  Most of the approaches fall into the following three groups: map-less reactive methods~\cite{target_rl}, memory-based methods~\cite{target_lstm1, target_lstm2, self_adaptive,transformer, transformer1,2021winner}, and explicit map-based methods~\cite{baseline, sscnav,occupation1, map_predict,pomp++,topological1, topological2}.

Among the Target Driven Navigation tasks, Object Goal Navigation~\cite{objectnav_defination} is the one that sets an object category as the target. The agent takes RGB-D images and poses as input, and output actions to search the target object. This task is similar to active visual search~\cite{avs} that has been studied for many years. A feasible solution is to formulate it as a POMDP problem. Some works model the belief by online learning~\cite{pomp++}, object-object/object-scene co-occur probability~\cite{avs_pomdp1, avs_pomdp4}, qualitative spatial relations~\cite{avs_pomdp3}. In comparison, we incorporate the knowledge of object relation into a distance prediction model. The model learns to predict the distance to the target object based on the explored semantic map, and a goal is selected according to the predicted distance using a designed strategy or a path planning method.

Recent works~\cite{baseline, attention, 2021winner, sscnav, transformer1} formulate this task as Reinforcement Learning (RL) and have achieved good performance. However, RL often suffers from low sample efficiency~\cite{sample_efficiency}. 
We replace the RL model in \cite{baseline} with a distance prediction model and a simple goal selection strategy.
The model can be trained offline by collecting semantic map samples and does not need to interact with the environment to get a reward.

Recent works also model scene priors by predicting elements out of sight to guide the agent to explore the target. 
Some works~\cite{sscnav, map_predict} use a semantic scene completion module to complete the unexplored scene.
Similar to it, some methods achieve image-level extrapolation~\cite{Im2Pano3D} or high-dimension feature space extrapolation~\cite{env_predict}. 
Accurately predicting the scene out of view is difficult as the exact relative pose between related objects is hard to determine.
We do not predict the exact semantics and their pose out of sight. Instead, we estimate the distance from the frontier area to the target, based on observed objects and spatial relations between objects.  The distance between the related objects is more likely to fall in a reasonable range, which is relatively easier to learn.



\section{APPROACH}
As shown in Fig.~\ref{fig: system_overview}, our method consists of three modules: a semantic mapping module, a target distance prediction module, and a local policy. The input of our system is the RGB-D image and the agent pose; the output is the next action. At each step, the RGB-D observation and the agent pose are used to update the bird's-eye view semantic map. Then, based on the semantic map and the learned knowledge, the distance prediction model estimates the shortest path length from the exploration frontier (the map cells between the explored and unexplored area) to the target. According to the distance map, the local policy selects a mid-term goal and gets the next action using a path planning method. 

Although the distance prediction may not be correct enough in the beginning, as the agent moves and receives more observation, the semantic map expands, the predicted distance map updates and becomes more accurate. With the update of the distance map, the agent could automatically and implicitly switch from random exploration to target searching and approaching, and reach the target object eventually. 

In Section~\ref{section:semantic_mapping} we briefly describe the Semantic Mapping module. Section~\ref{section:target_prediction} gives the definition of the target distance map and presents our model. In Section~\ref{section:local_policy}, we describe the details of the local policy. 


\begin{figure}[thpb]
    \centering
    \includegraphics[width=0.49\textwidth]{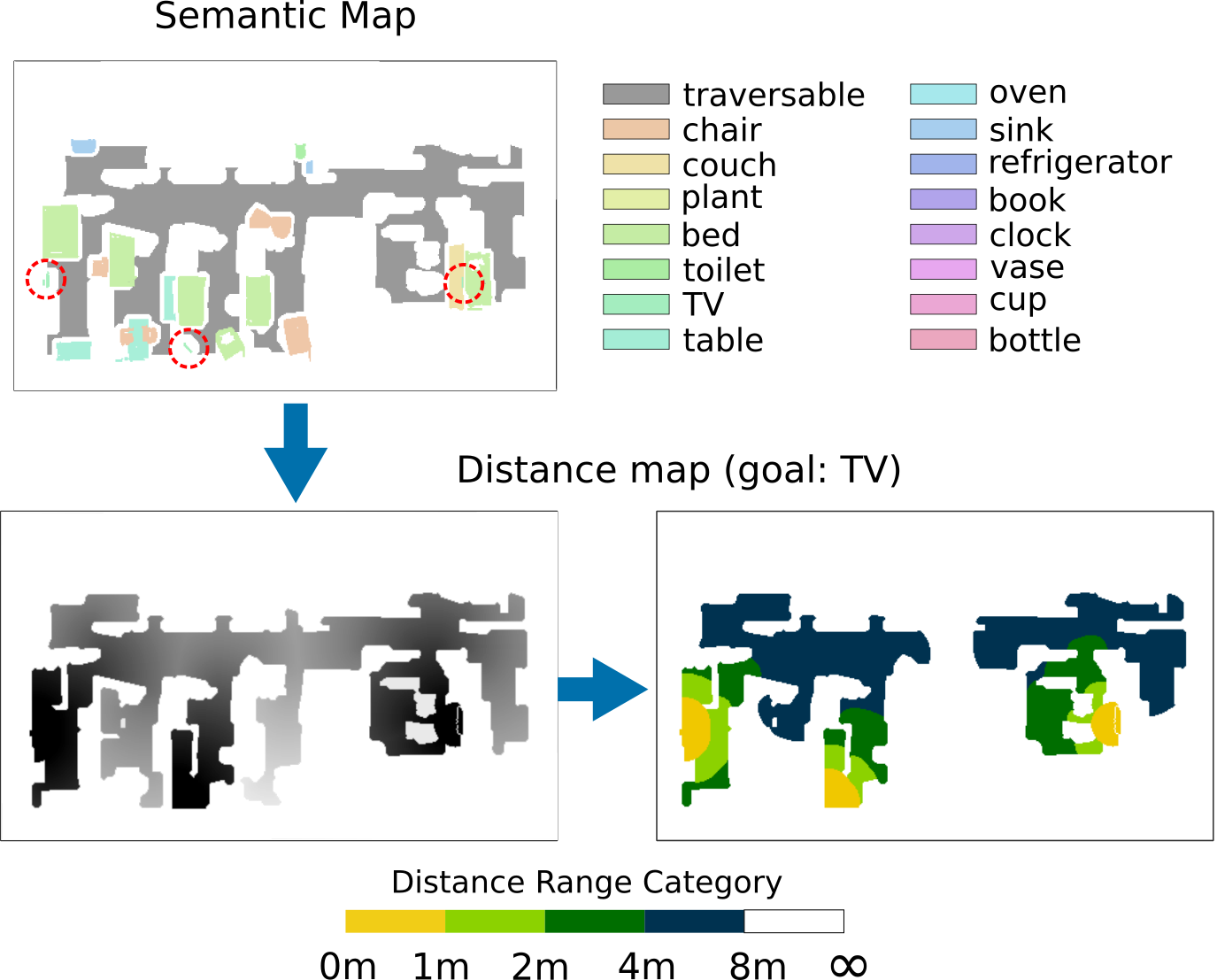}
    \caption{\textbf{Target Distance Ground Truth Map Example.} The first row is the semantic GT map; the second row is the continuous distance map (darker color corresponds to smaller distance value) and the discrete distance GT map for goal category ‘TV’. The red dotted circles indicate the location of ‘TV’.}
    \label{fig: data_example}
\vspace{-0.2cm}
\end{figure}
\subsection{Semantic Mapping}
\label{section:semantic_mapping}
We follow SemExp \cite{baseline} to build a bird's-eye view semantic map. The semantic map $L_S$ is with $k\times h\times w$, where $h\times w$ denotes the map size, $k = c_S + 2$, and $c_S$ is the total number of semantic categories. The other 2 channels represent obstacles and explored areas. In practice, we set the size of each cell in the semantic map to be 5cm $\times$ 5cm. 

Given the RGB-D image and the agent's pose, we use a semantic model, e.g. Mask-RCNN~\cite{mask_rcnn}, to predict semantic categories on the RGB image and then get a 3D semantic point cloud in the camera coordinate system using depth observation and camera intrinsics. The semantic point cloud is transformed to the global coordinate and projected to the ground to get the semantic map.


\subsection{Target Distance Prediction}
\label{section:target_prediction}
\textbf{Target Distance Map Definition:} The target distance map has the same map size and cell size as the semantic map. It stores the shortest path length the agent will travel to reach the target object. We define the distance value as zero for the cells inside the target object and infinite for the cells inside other obstacles except for those who contain the target (e.g., the table where a target object is on it).

The input of our target distance prediction model is the local semantic map $L_S$, which is $k \times 240 \times 240$. Our model is required to predict a local target distance map $L_D$ based on $L_S$ at each time step. 
In order to predict the distance to the target based on the explored semantic map, the model has to learn the spatial relations between different semantics (e.g., chairs often near a table). 
Since estimating the exact distance to the target is difficult, we formulate this problem as a classification problem instead of a regression problem. We split the distance into $n_b$ discrete bins, so each bin corresponds to a distance range category. 
In this paper, we set the number of distance range categories $n_b=5$, and the partition detail is shown in Fig.~\ref{fig: data_example}.

Our model is a fully convolutional neural network with 3 downsample ResBlocks~\cite{resblock}, 3 upsample ResBlocks, and concatenating low-level feature map and upsampled feature map on each level. The output is the local target distance map $L_D$ for the target. Although one can also choose to predict a $n_b$ channel distance map conditioned on the target category, in this paper, we set the output channel as $n_b \times n_T$, where $n_T$ is the number of target categories. In this way, every $n_b$ channels form a group, responsible for predicting the distance map of a specific target, so the distance prediction of all the target categories could be trained simultaneously.

\textbf{Training:}  As shown in Fig.~\ref{fig: data_example},  the ground-truth (GT) target distance map is generated based on the GT semantic map and traversable map. The distance values for the cells within the traversable areas are calculated and split into corresponding range categories. 
The local GT distance map is cropped from the global GT map according to the agent's pose. 
When collecting training data, the agent performs random exploration and collects the local semantic map as well as the local GT distance map. 
Note that there are no restrictions on how to collect the samples since the model is trained offline. 
\begin{figure}[thpb]
    \centering   
    \vspace{0.15cm}
    \subfigure[Integrating with Path Planning]{   
    \begin{minipage}[ab]{0.218\textwidth}
    \centering
    \includegraphics[width=1\textwidth]{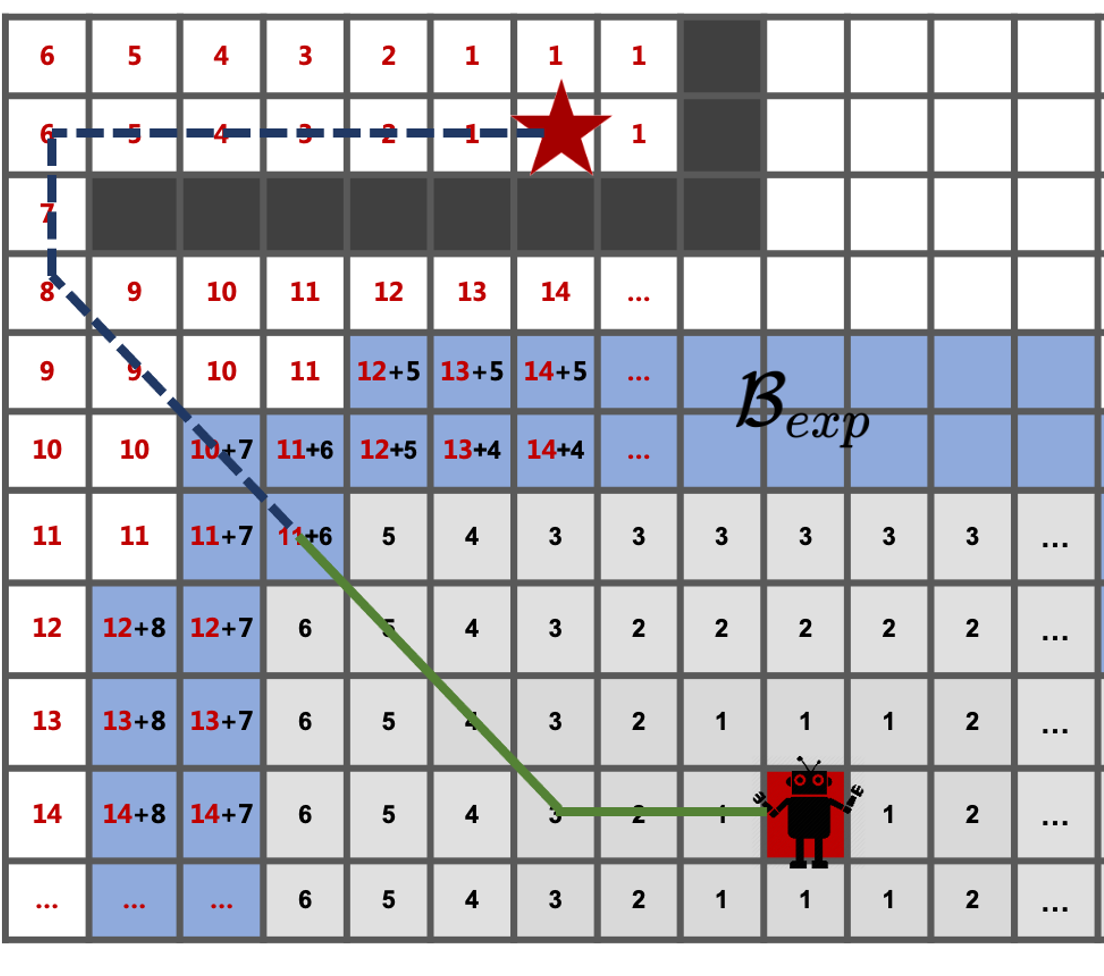} 
    \end{minipage}
    }
    \subfigure[Door-Exploring-First Strategy]{   
    \begin{minipage}[ab]{0.23\textwidth}
    \centering
    \includegraphics[width=1\textwidth]{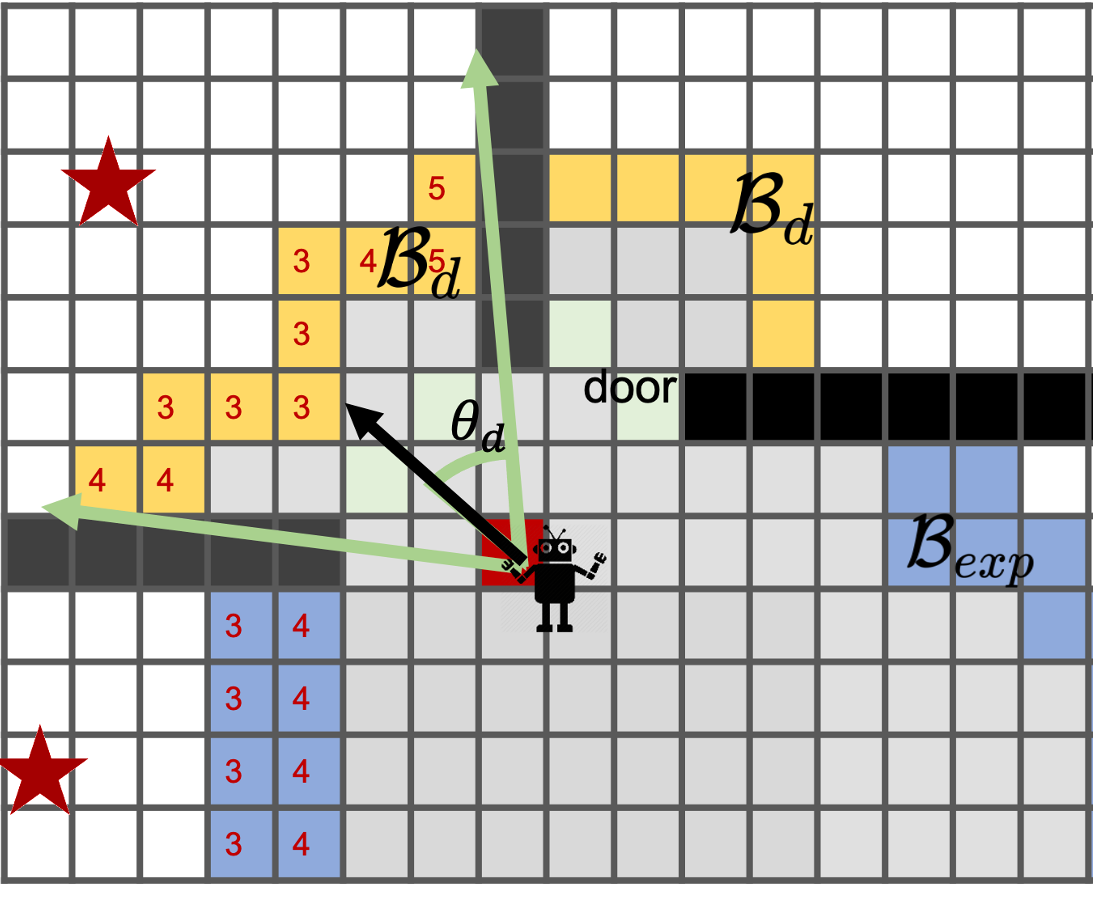}  
    \end{minipage}
    }

    \caption{\textbf{Terminology Illustration.} The red cells denote the current agent's position. 
    The grey cells denote the explored area, and the blue cells represent the frontier $\mathcal{B}_{exp}$ between the explored and unexplored area. 
    Black cells indicate obstacles. 
    \textbf{(a)}  Integrating Distance Prediction with Path Planning. The numbers within the frontier $\mathcal{B}_{exp}$ illustrate the target distance our model has to predict  (e.g., the length of the blue dotted line). Based on the predicted distance values within $\mathcal{B}_{exp}$, the path planning algorithm finds the optimal path (green line) to move closer to the target (the red star).
    \textbf{(b)} In the Door-Exploring-First strategy, if the agent observes a door or a passage (green cells), the intersection of a sector with angle $\theta_d$ and the frontier $\mathcal{B}_{exp}$ is defined as $\mathcal{B}_{d}$ (yellow cells).
    }
    \label{fig: boundary}
\vspace{-0.5cm}
\end{figure}
Besides, since the obstacles and semantics are already known in the explored area, if the target object is observed, the distance to the target can be obtained for cells within the explored area. 
If the target is out of view, the distance map within the explored area can also be calculated based on the distance value around the exploration frontier.
In this paper, we train our model using the pixel-wise Cross-Entropy loss only between the predicted distance range category and the ground-truth label within the area around the exploration frontier.

\subsection{Local Policy}
\label{section:local_policy}
Our local policy consists of two parts: a mid-term goal selection strategy and a path planner. At each time step, the goal selection strategy chooses mid-term goals on the local semantic map $L_S$ based on the local target distance map $L_{Dis}$. 
Here we design several simple strategies to demonstrate the effectiveness of the distance map. Following \cite{baseline}, we use the Fast Marching Method (FMM) \cite{FMM} algorithm to plan a path based on the obstacle channel of $L_S$. Finally, an action is selected according to the planned path. 

To obtain a mid-term goal, we design three strategies:
\subsubsection{\textbf{Integrating with Path Planning}}
With estimated distance to the target for the cells around the exploration frontiers, we plan a path with the smallest length from the current position to the target object. Given the distance map $L_D$ and the exploration frontier, we get the position of the mid-term goal 
$$
    p_{goal}=\mathop{\arg\min}_{p\in \mathcal{B}_{exp}} \{d(p_{agent},p)+L_{Dis}(p)\}, \eqno{(1)}
$$
where $p_{agent}$ is the current agent's position on the local map, $L_{Dis}(p)$ is the predicted distance value on the position $p$, $d(p_{agent},p)$ is the distance to the current position, which can be obtained by the path planning algorithm based on the obstacle map, $\mathcal{B}_{exp}$ is the area around the current exploration frontier (see Fig.~\ref{fig: boundary}). At each time step, the exploration area is expanded as the agent moves, so a new mid-term goal is selected on the new $\mathcal{B}_{exp}$ based on the updated $L_{Dis}(p)$. 
If all the predicted distance values in $\mathcal{B}_{exp}$ are infinite 
, the random exploration is adopted. If there is a target object on the semantic map (meaning that the target object is found), the area of the target object is selected as the mid-term goal.

\subsubsection{\textbf{Closest-First Strategy}} This strategy makes a little change to the above strategy by simply choosing the mid-term goal
$$
    p_{goal}=\mathop{\arg\min}_{p\in \mathcal{B}_{exp}} L_{Dis}(p), \eqno{(2)}
$$
which means we tend to move to the position where the predicted distance value is the smallest, regardless of the agent's current position.

\subsubsection{\textbf{Door-Exploring-First Strategy}}
When an agent faces a door or a passage leading to another room, it is more efficient to explore that room if the distance value in that room is smaller (which means the agent might see some objects related to the target through the opened door). Considering that, we design the door-exploring-first strategy based on the closest-first strategy.
The strategy first classifies whether a door or a passage is in the observed RGB image through a classification network based on ResNet50~\cite{resblock}. 
If true, we obtain the area that the door (passage) might lead to, that is, the area where the angle between the current orientation of the agent is less than $\theta_d$, as demonstrated in Fig.~\ref{fig: boundary}.
Then we obtain the intersection of this area and the frontier $\mathcal{B}_{exp}$, which is defined as $\mathcal{B}_{d}$.
Finally, We select the mid-term goal  
$$p_{goal}=\left\{
\begin{array}{cl}
\mathop{\arg\min}_{p\in \mathcal{B}_{d}} L_{Dis}(p) & {p_{door}\geq0.5, \ \mathcal{B}_{d} \neq \emptyset} \\
\mathop{\arg\min}_{p\in \mathcal{B}_{exp}} L_{Dis}(p) & {others}
\end{array} \right.
,$$
where $p_{door}$ means the probability of containing a door (passage). 
During training, we use the cross-entropy loss to train the door classifier. 
In the experiment, we find that when $\theta_d$ is small, the agent may change its heading direction after reaching the door, leading to walking back and forth at the same position. Therefore, in practice, we set $\theta_d$ as $\ang{120}$.

After generating a mid-term goal with one of the strategies, we use the FMM~\cite{FMM} algorithm to get the path, because the distance map used in this algorithm can be easily obtained based on the predicted distance map $L_{Dis}$. One can also use other path planning algorithms like $A^*$ \cite{a_star}. Note that if $L_{Dis}$ is accurately predicted, the ObjectNav task will become a traditional navigation task, since the goal coordinate can be viewed as known. 

\begin{table*}[h]
\caption{Object Goal Navigation Result (Success Rate$\uparrow$ / SPL$\uparrow$). -GT: using local GT semantic maps. -PP: integrating with Path Planning, -CF: Closest-First strategy, -DEF: Door-Exploring-First strategy. $^{\dagger}$ means our reimplementation.}
\vspace{-0.4cm}
\label{table:navigation_result}
\begin{center}
\begin{tabular}{l|c|c|c|c|c|c|c}
\hline
Method & Chair & Couch & Plant & Bed & Toilet & TV & Avg.\\
\hline

RandomExp-GT &  0.859/0.576 & 0.681/0.465 & \textbf{0.750}/0.489 & 0.579/0.400 & 0.790/0.422 & 0.887/0.579 & 0.762/0.496 \\

FrontierExp-GT~\cite{frontier_based_exploration} & 0.838/0.546 & 0.628/0.453 & 0.741/\textbf{0.519} & 0.497/0.334 & 0.790/0.448 & 0.877/0.680 & 0.734/0.493 \\

SemExp-GT~\cite{baseline} &\textbf{0.888}/0.627 & 0.730/0.518 & 0.585/0.409 & 0.597/0.411 & 0.790/0.476 & \textbf{0.910}/0.666 & 0.755/0.522 \\
SSCExp-GT~\cite{sscnav}$^{\dagger}$& 0.854/0.625 & 0.717/0.526 & 0.623/0.432 & 0.572/\textbf{0.415} & 0.783/0.484 & 0.899/\textbf{0.694} & 0.744/0.531\\
Ours-GT & 0.880/\textbf{0.687} & \textbf{0.735}/\textbf{0.590} & 0.637/0.478 & \textbf{0.610}/0.393 & \textbf{0.841}/\textbf{0.514} & 0.876/0.635 & \textbf{0.768}/\textbf{0.566} \\

\hline
RandomExp & 0.566/0.257 & 0.398/0.174 & 0.358/0.157 & 0.421/0.246 & 0.305/\textbf{0.158} & 0.000/0.000 & 0.403/0.193\\
FrontierExp~\cite{frontier_based_exploration} & 0.543/0.251 & 0.332/0.174 & 0.274/0.128 & 0.371/0.179 & 0.325/0.128 & 0.000/0.000 & 0.364/0.172\\
SemExp~\cite{baseline} & 0.622/0.289 & 0.385/0.220 & 0.344/0.141 & 0.415/0.213 & 0.280/0.124 & 0.000/0.000 & 0.410/0.197 \\
SSCExp~\cite{sscnav}$^{\dagger}$ & 0.552/0.305 & 0.319/0.183 & 0.382/\textbf{0.193} & 0.289/0.167 & 0.255/0.105 & 0.000/0.000 & 0.363/0.195\\ 
Ours-PP & \textbf{0.639}/0.290 & 0.389/0.179 & \textbf{0.387}/0.119 & \textbf{0.490}/\textbf{0.252} & 0.312/0.109 & \textbf{0.011}/\textbf{0.021} & \textbf{0.438}/0.190\\
Ours-CF & 0.611/\textbf{0.353} & 0.412/0.254 & \textbf{0.387}/0.156 & 0.421/0.238 & \textbf{0.338}/0.147 & \textbf{0.011}/0.002 & 0.428/0.231\\
Ours-DEF & 0.630/0.346 & \textbf{0.438}/\textbf{0.258} & 0.368/0.162 & 0.447/0.240 & 0.312/0.154 & \textbf{0.011}/0.003 & 0.436/\textbf{0.232}\\
\hline
\end{tabular}
\end{center}
\vspace{-0.5cm}
\end{table*}

\section{EXPERIMENTS}
\subsection{Experimental Setup}
\label{section:experimental_setup}
We perform experiments on the Habitat~\cite{habitat} platform with Matterport3D (MP3D)~\cite{mp3d} dataset. The training set consists of 54 scenes and the test set consists of 10 scenes. We follow~\cite{baseline} to set $n_T=6$ object goal categories: ‘chair’, ‘couch’, ‘plant’, ‘bed’, ‘toilet’, and ‘TV’. Same as \cite{baseline}, the semantic map has $c_S=15$ categories; the global semantic map size is $480 \times 480$ ($24m \times 24m$). We follow~\cite{baseline} to use a Mask-RCNN~\cite{mask_rcnn} pretrained on MS-COCO~\cite{coco} as the semantic model.

To collect training data, we randomly initialize the agent's position and perform random exploration mentioned in Section~\ref{section:experimental_setup}. We collect 1.2 million samples to train our model. 
We use pixel-wise Cross-Entropy loss within the area of 1m distance to the exploration frontiers. The loss weight for the distance range categories from 1m to infinite is set as from 5 to 1. 
Adam optimizer is used with a learning rate of 0.00001. 
During the evaluation, we split the scene into several floors according to the scene graph label of MP3D. For each scene, we first uniformly sample a floor, and then sample the goal among all the targets categories available on this floor. The agent is randomly initialized at the position with a distance margin to the target.
In this way, we sample a total of 1200 test episodes. The maximum length for each episode is 500 steps, and the success threshold is 1m. 

We use two metrics to evaluate the performance of ObjectNav: 
\begin{itemize}
    \item \textbf{Success Rate}: The ratio of the episode where the agent successfully reaches the goal;
    \item \textbf{SPL}~\cite{SPL_defination}: Success weighted by normalized inverse Path Length, which measures the efficiency of finding the goal. 
\end{itemize}


We compare our method with the following baselines:
\subsubsection{\textbf{Random Exploration (RandomExp)}} Instead of random walk, we design a simple strategy to urge the agent to explore the environment randomly. We set the mid-term goal as one of the corners of the local map. With the change of local map boundaries due to the agent's movement, the mid-term goal also changes with the boundaries. This goal is periodically switched clockwise among four corners per 100 steps. It also serves as a supplement to our goal select strategy as mentioned in Section~\ref{section:local_policy}.

\subsubsection{\textbf{Frontier-based Exploration (FrontierExp)}}
When performing Frontier-based Exploration~\cite{frontier_based_exploration}, the agent will select the nearest traversable and unvisited frontier cell as the mid-term goal. 

\subsubsection{\textbf{SemExp~\cite{baseline}}} SemExp consists of a semantic mapping module, an RL policy deciding mid-term goals based on the semantic map, and a local path planner based on FMM \cite{FMM}. 
The difference between it and our method is the way to select mid-term goals. 
Specifically, we utilize the target distance map rather than the RL policy. The RL model is trained with 10 million steps.

\subsubsection{\textbf{Semantic Scene Completion (SSCExp)}}
Following SSCNav \cite{sscnav}, we utilize 4 down-sampling residual blocks~\cite{resblock} and 5 up-sampling residual blocks to build the scene completion and confidence estimation module. 
It predicts the full semantic map and confidence map from the observed map constructed by the semantic mapping module.
We add this semantic scene completion module to SemExp~\cite{baseline}.
Then the RL policy generates a mid-term goal based on the completed maps. The scene completion and confidence estimation model is trained with 2.7 million samples and the RL model is trained with 10 million steps.



\subsection{Main Results}

Tab.~\ref{table:navigation_result} shows the result of our method. $\text{Ours-PP}$ indicates the strategy integrating with Path Planning, $\text{Ours-CF}$ is the Closest-First strategy, Ours-DEF corresponds to the Door-Exploring-First strategy. $*$-GT denotes using local ground-truth semantic maps within the explored area to replace the constructed map at each time step. $\text{Ours-GT}$ uses the Closest-First strategy to get action. Among the three strategies using distance map, Ours-PP has the highest success rate but low SPL; this is because the noise of the distance map makes the agent follow a zigzag route, increasing the total path length.
Ours-DEF has a good success rate and the best efficiency.

\textbf{Comparison with Baseline Methods.}
As shown in Tab.~\ref{table:navigation_result}, our method outperforms the baseline method SemExp\cite{baseline} (+2.6\% in success rate, +3.5\% in SPL). 
The result demonstrates that the target distance map is able to guide the agent to the target object more efficiently. Besides, our method only uses 12\% data (1.2 million vs. 10 million) to train the model compared with the RL-based method SemExp, indicating that our method is more sample efficient. 
It can be seen that all methods' success rates and SPL of target 'TV' are close to 0. Comparing with the performance using GT semantic maps, we attribute this phenomenon to the unsatisfying performance of the semantic model. As mentioned in \cite{pomp++}, the 3D reconstruction quality in some MP3D scenes is not gratifying. 
After eliminating the factor of semantic mapping by using GT semantic maps, the SPL of our method exceeds SemExp \cite{baseline} by 4.4\%. 
As for the baseline based on semantic scene completion, the performance of SSCExp is relatively poor in our experiment. Note that our setting is different from~\cite{sscnav} in camera angle, semantic map size, semantic categories, target categories, etc. We suscept the reason for the poor performance is that our local map size ($12m \times 12m$) is much larger than~\cite{sscnav} ($6m \times 6m$), causing it difficult to complete the scene in the local map.

\begin{figure}[thpb]
    \centering
    \vspace{0.05cm}
    \includegraphics[width=0.46\textwidth]{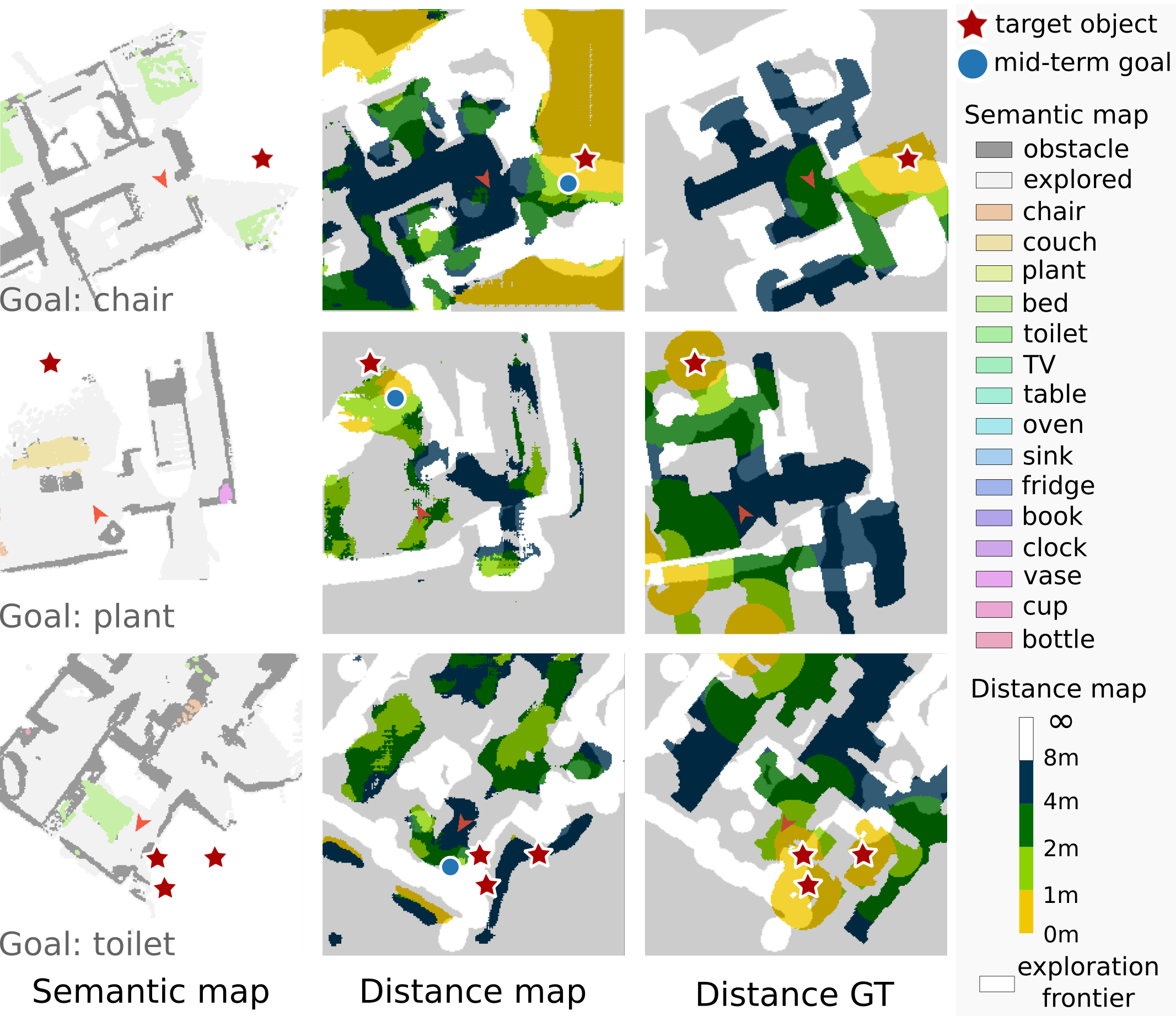}
    \caption{\textbf{Prediction Example.} Our model can guide the agent to the target object since the predicted directions are correct. From left to right are: local semantic maps, predicted local distance maps, and local distance GT maps. The red star denotes target objects. The blue dot corresponds to the mid-term goal. The red arrow denotes the agent's pose. The non-shaded area in the distance map indicates the area of exploration frontiers $\mathcal{B}_{exp}$.}
    \label{fig: prediction_example}
\end{figure}
\begin{figure}[thpb]
    \centering
    \includegraphics[width=0.48\textwidth]{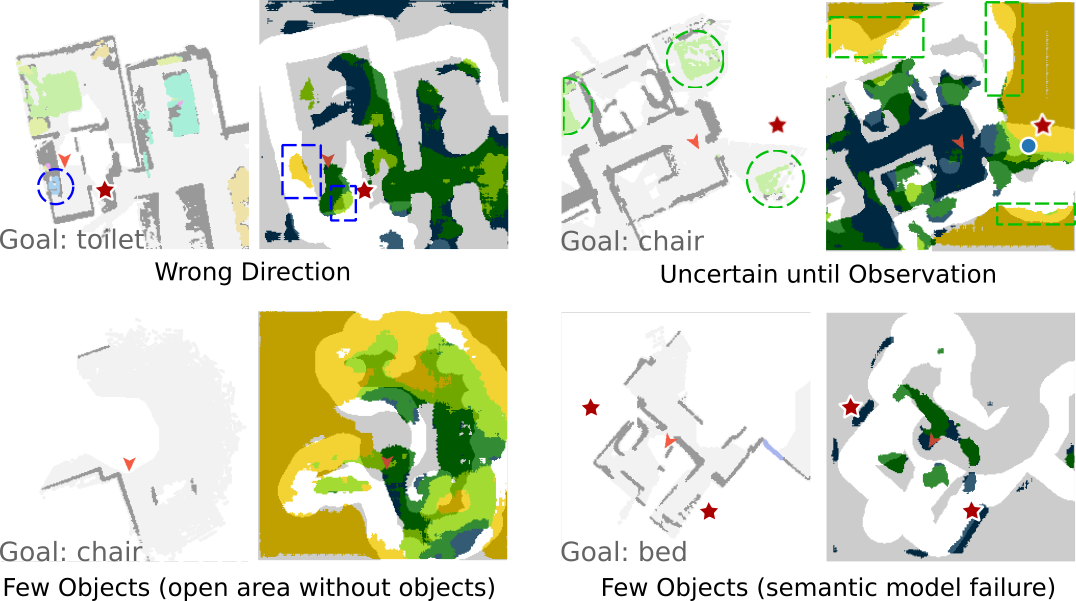}
    \caption{\textbf{Prediction Failure Cases Example.} The left images of each example are the semantic maps. The images on the right show the predicted distance map. \textbf{Upper left:} based on the observed object "sink" (blue dot circle), the direction prediction of the target "toilet" is wrong (blue dot box). \textbf{Upper right:} it also predicts the target "chair" has a "< 1m" distance (green dot box) around the other two beds (green dot circle), because it is uncertain whether there is a chair near a bed unless having explored all the area around the bed. \textbf{Bottom row:} our model cannot predict the distance due to the lack of semantic prediction (no object around the agent (bottom left), or the semantic model can not detect the object (bottom right)).
    }
    \label{fig: prediction_failurecase}
    \vspace{-0.2cm}
\end{figure}

\textbf{How does the target distance map work?} We test the quantitative performance of our model during the whole process of navigation. 
As shown in Tab.~\ref{table:dis_pred_result}, the performance is surprisingly low. 
Nevertheless, our model still guides the agent to the target object because the predicted directions to the target are correct (see Fig.~\ref{fig: prediction_example}). 
If it predicts a relatively small distance in the direction toward the target compared with other directions, the agent can still reach the target.

\begin{table}[h]
\vspace{0.05cm}
\caption{Performance of Distance Prediction Model}
\label{table:dis_pred_result}
\vspace{-0.4cm}
\begin{center}
\begin{tabular}{l|c|c|c|c|c|c}
\hline
Distance & <1m & 1-2m & 2-4m & 4-8m & >8m & Avg. \\
\hline
Precision & 0.045 & 0.193 & 0.078 & 0.109 & 0.844 & 0.254  \\
Recall & 0.354 & 0.081 & 0.064 & 0.047 & 0.842 & 0.278 \\
\hline
\end{tabular}
\end{center}
\vspace{-0.5cm}
\end{table}

\begin{table}[h]
\caption{Results of Different Representations }
\label{table:continous}
\vspace{-0.4cm}
\begin{center}
\begin{tabular}{l|c|c|c}
\hline
Representation & Partition (m) & Success Rate$\uparrow$ & SPL$\uparrow$  \\
\hline
\multirow{5}*{Discrete} & [1,2,4,8,$\infty$] & \textbf{0.428} & \textbf{0.231}  \\
& [1,2,4,$\infty$] & 0.416  & 0.226 \\
& [1,2,$\infty$] & 0.417 & \textbf{0.231}\\
& [1,$\infty$] &  0.405 & 0.226\\
& [2,4,8,12,$\infty$] & 0.390 & 0.209 \\
\hline
Continuous & - & 0.412 & 0.204   \\
\hline
\end{tabular}
\end{center}
\vspace{-0.4cm}
\end{table}

We further studied some wrong predictions. In Fig.~\ref{fig: prediction_failurecase} upper left, based on the observed object "sink" (blue dot circle), the direction of the target "toilet" is wrong (blue dot box).
In Fig.~\ref{fig: prediction_failurecase} upper right, it predicts the target "chair" has a "< 1m" distance (green dot box) around the other two beds (green dot circle), but in fact, there are no chairs around them. It demonstrates that our model has successfully learned the knowledge that "chairs may be close to beds." Nevertheless, unfortunately, it is still uncertain whether there is a chair near a bed unless the agent has explored all the area around the bed. On the contrary,  determining a target object is NOT near an unrelated object is much easier. 
This may, to some extent, explain why the performance for the categories "<= 8m" is poor, while the performance of the"> 8m" category in Tab.~\ref{table:dis_pred_result} is satisfying.  
Fig.~\ref{fig: prediction_failurecase} bottom row shows the cases that our model is not able to predict the distance due to the lack of semantic prediction. 
This often happens when there is no object around the agent (Fig.~\ref{fig: prediction_failurecase} bottom left), or the semantic model can not detect the object (Fig.~\ref{fig: prediction_failurecase} bottom right). This is also a reason for the low performance in Tab.~\ref{table:dis_pred_result}.  

\begin{figure}[thpb]
    \centering
    \vspace{0.05cm}
    \includegraphics[width=0.42\textwidth]{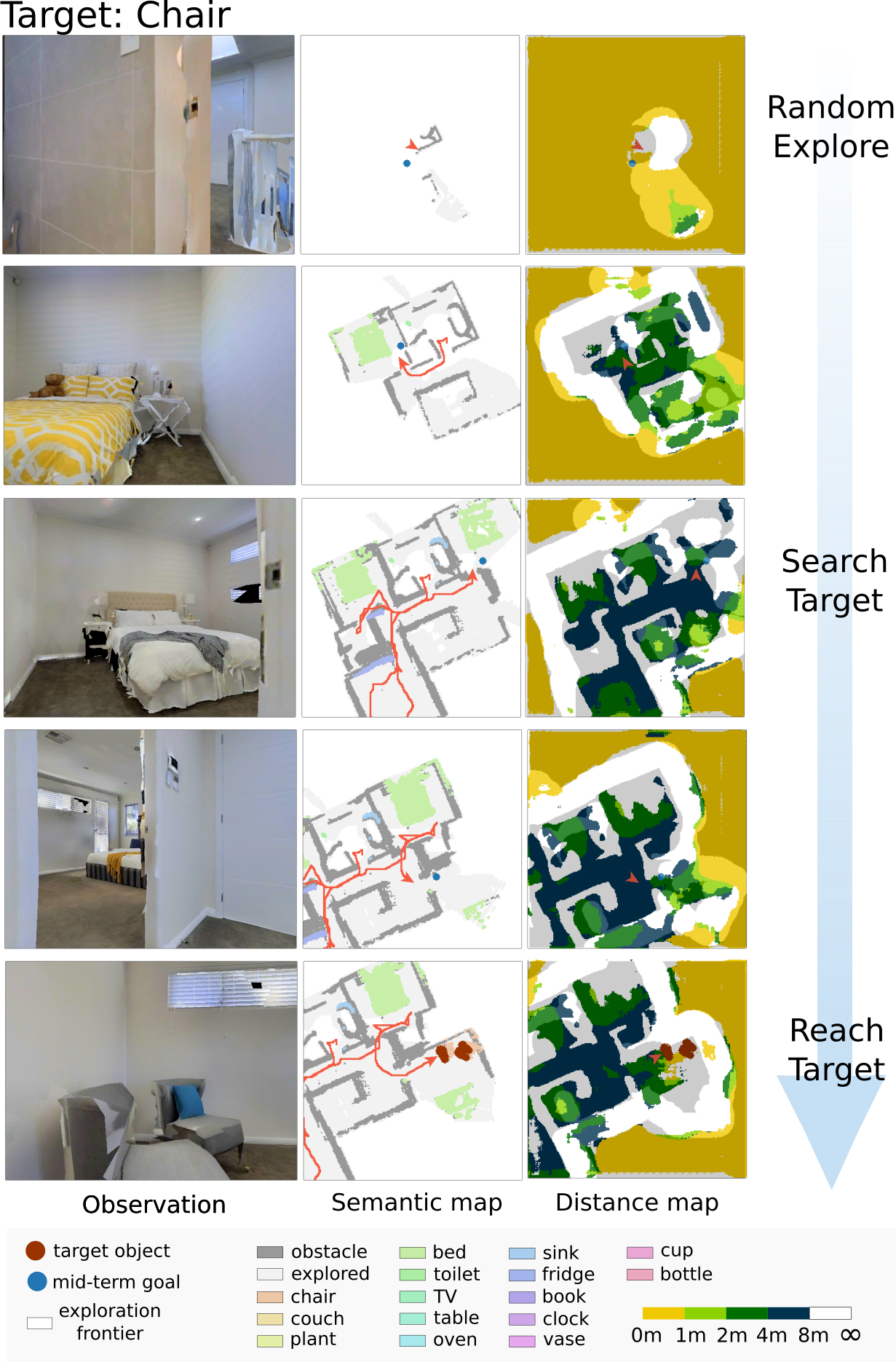}
    \caption{\textbf{Simulation episode example.} With the help of the target distance map, the agent first randomly explores, then searches the target around related objects (bed), and finally reaches the target (chair). From left to right are RGB observations, semantic map, and predicted distance map. The red arrow indicates the agent pose. The red line denotes the traversed path.}
    \label{fig: objectnav_example}
\vspace{-0.2cm}
\end{figure}

Fig.~\ref{fig: objectnav_example} illustrates how our method navigates to the target object with the help of the target distance map. In the beginning, the target distance prediction is random or invalid (Fig.~\ref{fig: objectnav_example} row 1), because there are few objects on the semantic map, or because the target is far from the agent. The agent can be seen as random exploration during this phase. As the agent explores and receives more observation, the model begins to predict the target distance distribution more accurately and guide the agent toward the direction of the potential target (in Fig.~\ref{fig: objectnav_example} row 2-3, the agent is looking for chairs around beds). If there is no target in the supposed direction, the distance map is corrected based on the new semantic map, and the agent will head to another direction with a low target distance. If the distance map is correct, the agent will reach the target (Fig.~\ref{fig: objectnav_example} row 4-5).


\textbf{Continuous or Discrete Representation for Target Distance Map.} In the target distance prediction model, we formulate it as the category classification problem. 
We also design a regression framework to predict the continuous distance.
Tab.~\ref{table:continous} shows the result of different representations using the Closest-First Strategy. 
The best result of discrete representation achieves a higher Success Rate and SPL than predicting the continuous distance. 
The result indicates that although the distance to a target is continuous, predicting a precise value is not easy. Besides, the result of different distance partitions indicates that the category of larger distance plays a less important role than smaller distance.


\begin{figure}[h]
    \centering
    \includegraphics[width=0.35\textwidth]{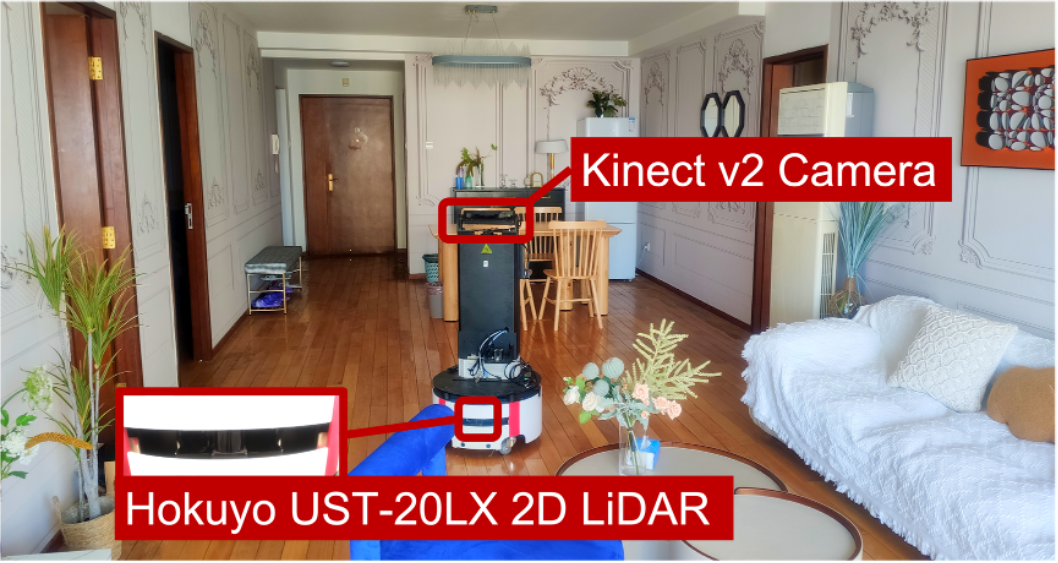}
    \caption{The real-world environment and the sensor configuration of our robot base. }
    \label{fig:robot_sensor}
\vspace{-0.2cm}
\end{figure}

\begin{figure}[thpb]
    \centering
    \vspace{0.15cm}
    \includegraphics[width=0.48\textwidth]{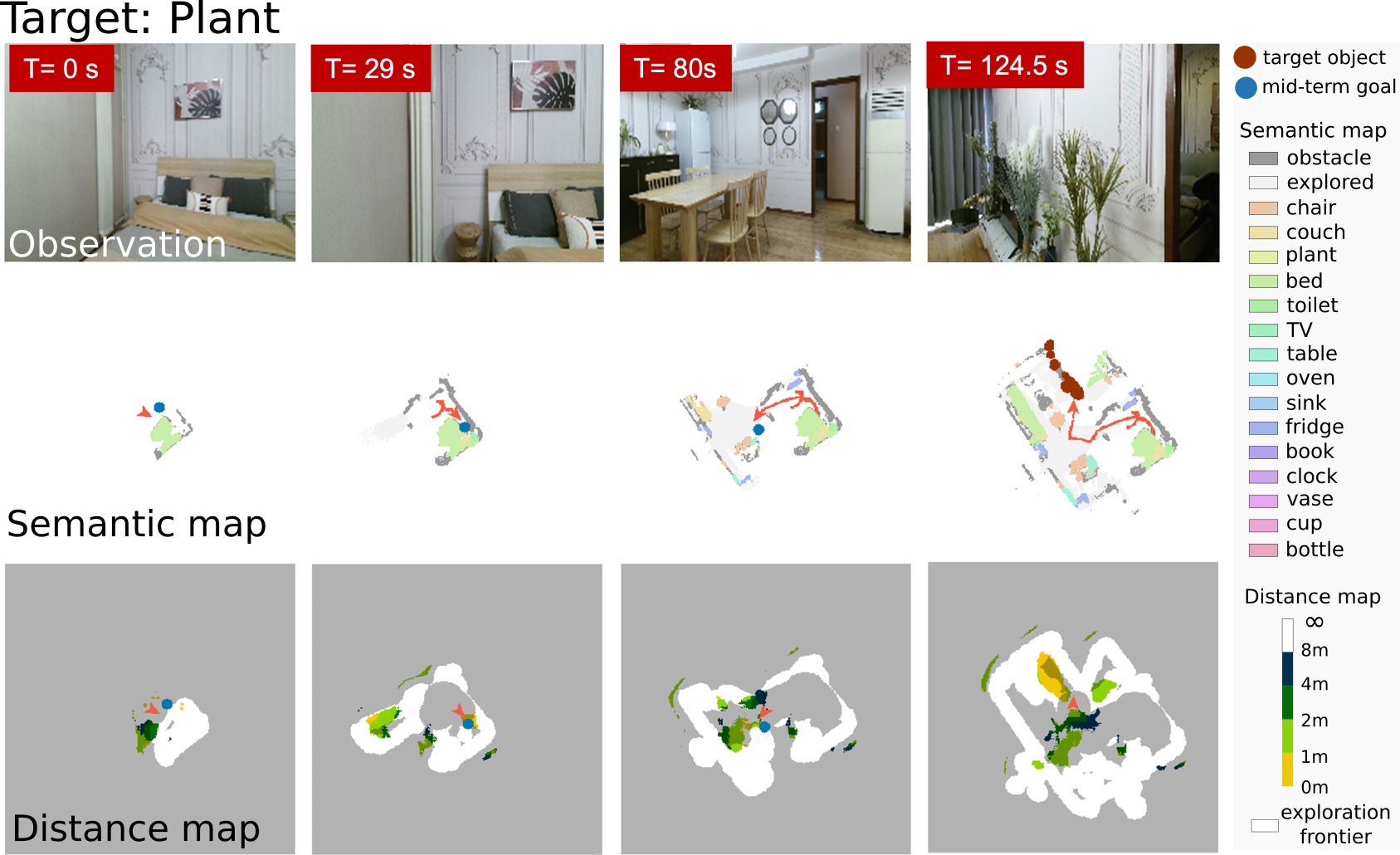}
    \caption{\textbf{Real-world episode example.} In a real-world episode with  "plant" as the target object, our method first searches the area around the bed, then the dining table. Finally, it searches the area around the TV stand and finds the target.}
    \label{fig: realworld_demo}
\vspace{-0.1cm}
\end{figure}

\subsection{Real World Experiment}
We also deploy our method to a real-world mobile base (see Fig.~\ref{fig:robot_sensor}). Our robot is equipped with a Kinect V2 camera, a 2D LiDAR, and an onboard computer (with an Intel i5-7500T CPU and an NVIDIA GeForce GTX 1060 GPU). The LiDAR is only used to perform localization, and no prior map is built. The average time consumption of each module is shown in Tab.~\ref{table:time_consumption}.
We test our method in a $120m^2$ house with one living room and three bedrooms. The max speed of our robot is limited to $0.2m/s$. The object goal categories in real-world experiments are the same as simulation experiments except "toilet." 
\begin{table}[!h]
\caption{Average time consumption of each module }
\label{table:time_consumption}
\vspace{-0.3cm}
\begin{center}
\begin{tabular}{p{1.18cm}|p{0.95cm}<{\centering}|p{1.05cm}<{\centering}|p{1.2cm}<{\centering}|p{1.05cm}<{\centering}|p{0.6cm}<{\centering}}
\hline
Module & Mask-RCNN  & Semantic Mapping  & Distance Prediction & Goal Selection & Total\\
\hline
Time (ms) & 276.7 & 31.9 & 2.7 & 54.6 & 365.9  \\
\hline
\end{tabular}
\end{center}
\vspace{-0.2cm}
\end{table}
\begin{table}[!h]
\caption{Real World Performance}
\label{table:realworld_performance}
\vspace{-0.3cm}
\begin{center}
\begin{tabular}{l|p{0.5cm}<{\centering}|p{0.6cm}<{\centering}|p{0.6cm}<{\centering}|p{0.6cm}<{\centering}|p{0.6cm}<{\centering}|p{0.6cm}<{\centering}}
\hline
 & Chair  & Couch  & Plant & Bed & TV & Avg.\\
\hline
Success Rate (\%) & 86.7 & 100 & 100 & 76.7 & 60.0&  84.7 \\
\hline
Time (s) & 140.7 & 129.6 & 100.5 & 126.6 & 176.4 & 134.8 \\
\hline
\end{tabular}
\end{center}
\vspace{-0.2cm}
\end{table}
 The goal "toilet" can not be reached because there are door saddles that are not traversable for our robot base. Fig.~\ref{fig: realworld_demo} shows one of the episodes with target "plant." In a total of 150 episodes, we evaluate the success rate and the average time to reach the goal among success episodes. The quantitative result in Tab.~\ref{table:realworld_performance} demonstrates that our method is transferred well to the real world. The real-world performance is much higher than that in simulation due to the better performance of the semantic model in the real world.

\subsection{Failure Cases}
\label{section:failure_cases}
In the experiments, we find that most failure cases are due to low semantic map accuracy. Sometimes the semantic model can not detect the object, and sometimes there is wrong detection or wrong projection to the ground due to the semantic segmentation noise or depth image noise. 
Besides, since the camera angle is fixed, and has a certain height from the ground, sometimes it cannot see the object in a small room, like the object "toilet" in a lavatory.
We also find some cases similar to the failure modes ``Goal Bug'' and ``Void'' as mentioned in \cite{2021winner}.

\section{CONCLUSIONS}
This paper presents a navigation framework based on predicting the distance to the target object. In detail, we design a model which takes a bird's-eye view semantic map as input, and estimates the path length from frontier cells to the target. 
Based on the distance map, the agent could navigate to the target objects with simple goal selection strategies and a path planning algorithm. 
Experimental results on the MP3D dataset demonstrate that our method outperforms baselines methods on success rate and SPL. The real-robot experiment also shows that our method can be transferred to the real world.
Future work would focus on predicting the target distance map more accurately, such as using the room-type prediction as auxiliary tasks.
We believe that with a more powerful target prediction model and RL policy, our method will achieve much better performance. 








\bibliographystyle{unsrt}

\end{document}